\documentclass[10pt, a4paper]{article}
\usepackage{lrec}
\usepackage{multibib}
\newcites{languageresource}{Language Resources}
\usepackage{graphicx}
\usepackage{tabularx}
\usepackage{soul}
\usepackage{titlesec}
\titleformat{\section}{\normalfont\large\bf\center}{\thesection.}{1em}{}
\titleformat{\subsection}{\normalfont\SmallTitleFont\bf\raggedright}{\thesubsection.}{1em}{}
\titleformat{\subsubsection}{\normalfont\normalsize\bf\raggedright}{\thesubsubsection.}{1em}{}
\renewcommand\thesection{\arabic{section}}
\renewcommand\thesubsection{\thesection.\arabic{subsection}}
\renewcommand\thesubsubsection{\thesubsection.\arabic{subsubsection}}

\usepackage{epstopdf}
\usepackage[utf8]{inputenc}

\usepackage{hyperref}
\usepackage{xstring}

\usepackage{color}

\usepackage{listings}
\usepackage{xcolor}

\colorlet{punct}{red!60!black}
\definecolor{background}{HTML}{EEEEEE}
\definecolor{delim}{RGB}{20,105,176}
\colorlet{numb}{magenta!60!black}

\lstdefinelanguage{json}{
    basicstyle=\normalfont\ttfamily,
    numbers=left,
    numberstyle=\scriptsize,
    stepnumber=1,
    numbersep=8pt,
    showstringspaces=false,
    breaklines=true,
    frame=lines,
    backgroundcolor=\color{background},
    literate=
     *{0}{{{\color{numb}0}}}{1}
      {1}{{{\color{numb}1}}}{1}
      {2}{{{\color{numb}2}}}{1}
      {3}{{{\color{numb}3}}}{1}
      {4}{{{\color{numb}4}}}{1}
      {5}{{{\color{numb}5}}}{1}
      {6}{{{\color{numb}6}}}{1}
      {7}{{{\color{numb}7}}}{1}
      {8}{{{\color{numb}8}}}{1}
      {9}{{{\color{numb}9}}}{1}
      {:}{{{\color{punct}{:}}}}{1}
      {,}{{{\color{punct}{,}}}}{1}
      {\{}{{{\color{delim}{\{}}}}{1}
      {\}}{{{\color{delim}{\}}}}}{1}
      {[}{{{\color{delim}{[}}}}{1}
      {]}{{{\color{delim}{]}}}}{1},
}
\title{Automated Extraction of Socio-political Events from News (AESPEN): Workshop and Shared Task Report}

\name{Ali Hürriyetoğlu$^{\ast}$, Vanni Zavarella$^{\dagger}$, Hristo Tanev$^{\dagger}$, Erdem Yörük$^{\ast}$, Ali Safaya$^{\ast}$, Osman Mutlu$^{\ast}$}

\address{$^{\ast}$Koç University, \\
          Rumelifeneri Yolu 34450, Sarıyer, İstanbul/Turkey \\
          \{ahurriyetoglu, eryoruk, asafaya19, omutlu\}@ku.edu.tr \\ \\
        $^{\dagger}$European Commission Joint Research Centre \\
         Via E. Fermi, 2749, 21027 Ispra VA, Italy \\
         \{vanni.zavarella, hristo.tanev\}@ec.europa.eu\\}

\abstract{
We describe our effort on automated extraction of socio-political events from news in the scope of a workshop and a shared task we organized at Language Resources and Evaluation Conference (LREC 2020).
We believe the event extraction studies in computational linguistics and social and political sciences should further support each other in order to enable large scale socio-political event information collection across sources, countries, and languages.
The event consists of regular research papers and a shared task, which is about event sentence coreference identification (ESCI), tracks. All submissions were reviewed by five members of the program committee. 
The workshop attracted research papers related to evaluation of machine learning methodologies, language resources, material conflict forecasting, and a shared task participation report in the scope of socio-political event information collection.
It has shown us the volume and variety of both the data sources and event information collection approaches related to socio-political events and the need to fill the gap between automated text processing techniques and requirements of social and political sciences. \\ \newline }

\begin{document}

\maketitleabstract


\section{Introduction}

Automatic construction of socio-political event databases has long been a challenge for the natural language processing (NLP) and social and political science communities in terms of algorithmic approaches and language resources required to develop automated tools~\cite{Chenoweth+13,Weidman+19,Raleigh+10}. At the same time, social and political scientists have been working on creating socio-political event databases for decades using manual ~\cite{Yoruk12}, semi-automatic~\cite{Nardulli+15}, and automatic approaches ~\cite{Leetaru+13,Boschee+13,Schrodt+14,Sonmez+16}. However, the results yielded by these approaches to date are either not of sufficient quality or require tremendous effort to be replicated on new data \cite{Wang+16,Ward+13,Ettinger+17}. On the one hand, manual or semi-automatic methods require high-quality human effort; on the other hand, state-of-the-art automated event detection systems are not accurate enough for their output to be used directly without human moderation.

The NLP community has achieved some consensus on the treatment of events both in terms of task definition and appropriate techniques for their detection~\cite{Pustejovsky+05,Doddington+2004,Song+15,Getman+18}. However, in order to be useful, these formalisms and related systems need to be adjusted or extended for each type of event in relation to certain use cases. The social and political scientists spend a similar effort for formalising event types such as (CAMEO)~\cite{Gerner+02} and implement the aforementioned systems that vary from rule-based to fully automatic approaches. Unfortunately, any new project in this line still finds itself making design decisions such as using only the heading sentences in a news article or not considering coreference information~\cite{Boschee+13} without being able to quantify their effect. Therefore, we think these communities should investigate ways of supporting each other in order to reach a consensus and enable any prospective event information collection project as robustly and predictably as possible.

Given the aforementioned limitations, there is an increasing tendency to rely on machine learning (ML) and NLP methods to deal better with the vast amount and variety of data to be processed. Consequently, we thought it was time to held a workshop on Automated Extraction of Socio-political Events from News (AESPEN)\footnote{\url{https://emw.ku.edu.tr/aespen-2020/}, accessed on April 18, 2020.} at Language Resources and Evaluation Conference (LREC 2020).\footnote{\url{https://lrec2020.lrec-conf.org/}, accessed on April 18, 2020.} The purpose of this workshop was to inspire the emergence of innovative technological and scientific solutions in the field of event detection and event metadata extraction from news, as well as the development of evaluation metrics for socio-political event recognition. Moreover, the workshop aimed at triggering a deeper understanding of the usability of socio-political event datasets.

We organized a shared task as a continuation of the Conference and Labs of the Evaluation Forum (CLEF 2019) task ProtestNews~\cite{Hurriyetoglu+19a,Hurriyetoglu+19b}, which was on cross-context document classification, event sentence detection, and event extraction pertaining to protest events. We aimed at establishing a benchmark for the event sentence coreference identification (ESCI) sub-task within the scope of the AESPEN workshop. The scope of this shared task was on clustering given event related sentences so that each cluster consists of sentences about the same event.

We provide details of our motivation in Section~\ref{motivate}. Then, we introduce the ESCI shared task in Section~\ref{sharedtask}. Finally, we briefly describe the accepted papers and the shared task results in Section~\ref{submissions}. We conclude this report in section~\ref{conclusion}.

\section{Motivation}
\label{motivate}

Automating political event collection requires the availability of gold-standard corpora that can be used for system development and evaluation. Moreover, automated tool performances need to be reproducible and comparable. Although a tremendous effort is being spent on creating socio-political event databases such as ACLED~\cite{Raleigh+10}, the Global Database of Events, Language, and Tone (GDELT)~\cite{Leetaru+13}, the Mass Mobilization on Autocracies Database (MMAD)~\cite{Weidman+19}, the Integrated Crisis Early Warning System (ICEWS)~\cite{Boschee+13}, and the Protest Dataset 30 European countries (PolDem)~\cite{Kriesi+19} we believe there is still a lot of room for improvement and harmonisation of the event schemas and tasks. This limitation causes the definition of the events and automated event information collection tool performances to be restricted to single projects. Consequently, the lack of comparable and reproducible settings hinders progress on this task.

We invited contributions from researchers in NLP, ML and Artificial Intelligence (AI) involved in automated event data collection, as well as researchers in social and political sciences, conflict analysis and peace studies, who make use of this kind of data for their analytical work. Our goal was to enable the emergence of innovative NLP and information extraction (IE) solutions that can deal with the current stream of information, manage the risks of information overload, identify different sources and perspectives, and provide unitary and intelligible representations of the larger and long-term storylines behind news articles.

Our workshop provided a venue for discussing the creation and facilitation of language resources in the social and political sciences domain. Social and political scientists were interested in reporting and discussing the automated tools and comparing traditional coding approaches with automated tools. Computational linguistics and machine learning practitioners and researchers benefited from being challenged by real-world use cases, in terms of event data extraction, representation and aggregation.

We invited work on all aspects of automated coding of socio-political events from monolingual or multilingual news sources. This includes (but is not limited to) the following topics: event metadata extraction, source bias mitigation, event data schema and representation, event information duplication detection, extracting events beyond a sentence in a document, training data collection and annotation processes, event coreference (in- and cross-document), sub-event and event subset relations, event dataset evaluation and validity metrics, event datasets quality assessments, defining, populating and facilitating event ontologies, automated tools for relevant subtasks, understanding the limits that are introduced by copyright rules and ethical concerns and ethical design.

\section{Shared Task}
\label{sharedtask}

A news article may contain one or more events that are expressed with one or more sentences. Identifying event sentences that are about the same event is necessary in order to collect event information robustly. Therefore, we should develop methods that are able to identify whether a group of sentences are about the same event. Reliable identification of this relation will enable us to determine how many events are reported in a news article as well. Moreover, solving this problem has the potential to facilitate cross-document event sentence relation identification in the long term. Therefore, we should develop methods that are able to identify whether a group of event sentences are about the same event. Consequently, we organized the ESCI shared task in the hopes of attracting attention to this problem and possibly provide a benchmark for it.
 
We examined our gold standard corpus that contains 1,290 events in 712 documents annotated at token level for their event information~\cite{Hurriyetoglu+19a,Hurriyetoglu+19b,Hurriyetoglu+20}. These documents are the positively labelled instances of random samples and active learning based samples based on these random samples. We have observed that 60\% of the news articles contain information about a single event. The remaining documents contain information about multiple events, which sums up to 45\% of the total event count. Only 45\% of the events are expressed with only a single sentence.

Consequently, we think protest event collection systems should take these phenomena into account and introduce the ESCI shared task. As training data participants of the data challenge received event related sentences and their true clustering in a news article, in which a cluster represents all sentences about an event. This data was extracted from 404 documents. The documents that contain a single event sentence were excluded from this exercise, since there is only one possible clustering in that case. The number of events per document in the training data is 1 for 207 and 2 for 132 documents. The remaining 65 documents contain 3 or more events. The task of the participants was to develop systems that can predict grouping of the given sentences that consists of events on test data, extracted from 100 documents, and that was delivered to them one week before the deadline. The correct grouping of the test set was not shared with the participants. The evaluation metric is Adjusted Rand Index (ARI) as implemented by Scikit-learn~\cite{Hubert85}.\footnote{\url{https://scikit-learn.org/stable/modules/generated/sklearn.metrics.adjusted_rand_score.html}, accessed on April 19, 2020.} We calculated macro and micro versions of this score. The macro version calculates average of the per document scores from all of the documents independent of how many event sentences are there in each document. However, the micro score weights the per document score with the number of the event sentences in a document. We report the F1 score that is calculated similarly as well.

The event type is a protest in the scope of this task. The event we simply refer to as protest events are comprised within the scope of contentious politics. Contentious politics events refers to politically motivated collective action events which lay outside the official mechanisms of political participation associated with formal government institutions of the country in which the said action takes place.\footnote{You can find detailed information about how a protest is defined and how event sentences are labelled on our annotation manual, which is on \url{https://github.com/emerging-welfare/general_info/tree/master/annotation-manuals}.}

The data is shared with the researchers, who signed an application form that limits the use of the data only for research purposes, as a file that contains lines of JSON objects. Each JSON object contain all event sentences that are identified in a news article and their clustering, which is found in the \textit{event\_clusters} field.

A sample JSON object is presented below. The \textit{url} field is provided only as an ID. The numbers in the \textit{sentence\_no} correspond to sentences in the \textit{sentences} field in the same order. The \textit{event\_clusters} field provides the correct clustering of the event sentences. For instance, below in Listing 1, the first and third sentences are about the same event. But, the second sentence is about a separate event.

\begin{lstlisting}[language=json,firstnumber=1,caption=Event sentences that are extracted from a document in the order they occur in a sentence.]
{
  "url": "http://www.newindianexpress.com/nation/2009/aug/25/congress-demands-advanis-apology-80257",
  "sentences": [
    "Singh had recently blamed Advani for coming to Gujarat Chief Minister Narendra Modi ' s rescue and ensured that he was not sacked , in the wake of the riots .", 
    "On Kandahar plane hijack issue , Singh said Advani was not speaking the truth .",
    "Elaborating on the three issues , Singhvi said , The BJP gave sermons on Raj Dharma and turned a Nelson ' s eye to the communal carnage , which became a big blot on the fair name of the country ."
  ],
  "sentence_no": [4, 6, 14],
  "event_clusters": [[4, 14], [6]]
}
\end{lstlisting}

We have calculated three baseline scores on the test data. First, we checked score of a dummy predictor that assigns all event sentences to a single cluster all the time, i.e., minimum cluster prediction (MinC). Second, another dummy baseline predicts as each event sentence as being in a separate cluster in a document, i.e., maximum cluster prediction (MaxC). Finally, we used BERT sentence representations~\cite{Devlin+19} to train a multilayer perceptron (MLP) model that i) first evaluates each possible sentence pair in the document, ii) then assign a positive or negative label indicating that this pair of sentences is co-referent, iii) finally using the correlation clustering algorithm~\cite{Bansal+04} we take those labeled pairs and cluster them.~\footnote{The code for this system is available on \url{https://github.com/alisafaya/event-coreference}, accessed on April 21, 2020.} The scores of these methods are provided in Table \ref{table:baselines} as \textit{MinC}, \textit{MaxC}, and \textit{MLP}. The slightly low scores obtained from the dummy systems direct us to use the MLP system as the baseline we share with the participants. Note that the strength of the dummy baselines changes according to data distribution in test data.

\begin{table}[!h]
\centering
\begin{tabular}{|c|c|c|c|c|}

      \hline
      & \multicolumn{2}{c|}{ARI} & \multicolumn{2}{c|}{F1} \\
      \hline
       & Macro & Micro & Macro & Micro \\
      \hline
      MinC & .5000	 & .4040 & .5000 & .4040 \\ 
      \hline
      MaxC & .1071 & .0628 & .3476 	 & .3722 \\ 
      \hline
      MLP & .5077 & .4064 & .5560 & .4840 \\ 
      \hline

\end{tabular}
\caption{Adjusted Random Index (ARI) and F1 for each baseline system.}
\label{table:baselines}
\end{table}

\section{Submissions}
\label{submissions}

The workshop has attracted nine papers as regular paper submissions and one as a shared task participation report. The shared task report and seven of the regular papers were accepted on the basis of the reviews, which were five per paper, performed by the program committee members. 
 
The accepted regular papers can be grouped as i) evaluation of state-of-the-art machine learning approaches by \newcite{Buyukoz+20}, \newcite{Olsson+20}, and \newcite{Piskorski+20}, ii) introduction of a new data set by \newcite{Radford20}, iii) projects of event information collection by \newcite{Osorio+20} and \newcite{Papanikolaou+20}, and iv) forecasting of political conflict by \newcite{Halkia+20}. 

The evaluation of \newcite{Buyukoz+20} and \newcite{Olsson+20} show that state-of-the-art deep learning models such as BERT and ELMo~\cite{Peters+18} yield consistently higher performance than traditional ML methods such as support vector machines (SVM) on conflict and protest event data respectively. \newcite{Piskorski+20} have found that TF-IDF weighted character n-gram based SVM model performs better than an SVM model that uses pre-trained embeddings such as GLOVE~\cite{Pennington+14}, BERT, and FASTTEXT~\cite{Mikolov+18} in most of the experiments on conflict data.

\newcite{Radford20} introduces the dataset \textit{Headlines of War} for cross-document coreference resolution for the news headlines. The dataset consists of positive samples from \textit{Militarized Interstate Disputes} dataset and negative samples from New York Times.\footnote{\url{https://spiderbites.nytimes.com}, accessed on April 21, 2020.} The description of this invaluable resource is accompanied with a detailed discussion of its utility and caveats.

\newcite{Osorio+20} introduce Hadath that is a supervised protocol for event information collection from Arabic sources. The utility of Hadath was demonstrated in processing news reported between 2012 and 2012 in Afghanistan. In the scope of the other event information collection study, \newcite{Papanikolaou+20} processed two news sources in Greek from Greece to create a database of protest events for the period between 1996 and 2014. Osorio et al and Papanikolaou and Papageorgiou utilized fully automatic tools that integrate supervised machine learning and rule based methodologies at various degrees.

Finally, \newcite{Halkia+20} presents a material conflict forecasting study that exploits available event databases GDELT and ICEWS. Their results demonstrate that it is possible to correctly predict social upheaval using the methodology they propose, which utilizes Long-Short Term Memory (LSTM)~\cite{Hochreiter+97}.


We have received expression of interest from 12 research teams, of which 6 teams signed the application form and received the data. Two of these teams sent their predictions on the test data. The scores of these methods are illustrated in Table~\ref{table:results}. Finally, only \newcite{Ors+20} submitted a paper about their work. This team reported their work as consisting of three steps. First, they use a transformer based model, which is ALBERT~\cite{Lan+20}, to predict whether a pair of sentences refer to the same event or not. Later, they use these predictions as the initial scores and recalculate the pair scores by considering the relation of sentences in a pair with respect to other sentences. As the last step, final scores between these sentences are used to construct the clusters, starting with the pairs with the highest scores.

\begin{table}[!h]
\centering
\begin{tabular}{|c|c|c|c|c|}

      \hline
      & \multicolumn{2}{c|}{ARI} & \multicolumn{2}{c|}{F1} \\
      \hline
       &  Macro & Micro & Macro & Micro \\
      \hline
      Örs et al. & .6006 & .4644 & .6736 & .5898 \\ 
      \hline
      UNC Charlotte & .3388 & .3253 & .4352 & .3284 \\ 
      \hline

\end{tabular}
\caption{Adjusted Random Index (ARI) and F1 for each baseline system.}
\label{table:results}
\end{table}

\section{Concluding Remarks}
\label{conclusion}

We have provided a brief summary of the workshop Automated Extraction of Socio-political Events from News (AESPEN) and the shared task Event Sentence Coreference Identification (ESCI) we organized in the scope of Language Resources and Evaluation Conference (LREC 2020). The variety of the submitted papers show that we could bring the ML, NLP, and social and political science communities together. Although the breadth of the topics were limited, the technical depth and timeliness of the contributions show that the workshop contribute to the discipline of automatic extraction of socio-political events. The papers about processing Arabic and Greek sources are significant contributions to the understanding of how should we handle languages other than English. Finally, the shared task ESCI demonstrated the prevalence of the event coreferences, some baselines for handling them, and a state-of-the-art system that is able to tackle this task.

We consider this workshop as a beginning. We expect this effort to be extended both in depth and in breadth since we think the work presented is only the tip of the iceberg considering the recent projects and technical potential introduced by deep learning technologies.

\section{Acknowledgements}
The authors from Koç University are funded by the European Research Council (ERC) Starting Grant 714868 awarded to Dr. Erdem Y\"{o}r\"{u}k for his project Emerging Welfare. We appreciate contributions of the program committee members, who are in alphabetical order Svetla Boycheva, Fırat Duruşan, Theresa Gessler, Christian Göbel, Burak Gürel, Matina Halkia, Sophia Hunger, J. Craig Jenkins, Liron Lavi, Jasmine Lorenzini, Bernardo Magnini, Osman Mutlu, Nelleke Oostdijk, Arzucan Özgür, Jakub Piskorski, Lidia Pivovarova, Benjamin J. Radford, Clionadh Raleigh, Ali Safaya, Parang Saraf, Philip Schrodt, Manuela Speranza, Aline Villavicencio, Çağrı Yoltar, Kalliopi Zervanou and of the keynote speaker Clionadh Raleigh. We are grateful to the management of the Competence Centre on Text Mining and Analysis (CC-TMA) at European Commission Joint Research Center (JRC) for the support. Any opinions, findings, conclusions, or suggestions expressed here are those of the authors and do not necessarily reflect the view of the sponsor(s) or authors' employer(s).

\section{Bibliographical References}\label{reference}

\bibliographystyle{lrec}
\bibliography{emwwp2}


\end{document}